\documentclass[10pt,twocolumn,letterpaper]{article}

\usepackage{cvpr}
\usepackage{times}
\usepackage{epsfig}
\usepackage{graphicx}
\usepackage{amsmath}
\usepackage{amssymb}
\usepackage[ruled]{algorithm2e} 
\usepackage{bm}
\usepackage{multirow}
\usepackage{float}

\usepackage[pagebackref=true,breaklinks=true,letterpaper=true,colorlinks,bookmarks=false]{hyperref}

\cvprfinalcopy 

\begin{document}

\title{Beyond Part Models: Person Retrieval with Refined Part Pooling\\ (and A Strong Convolutional Baseline)}
\author{Yifan Sun$^{\dag}$,\ Liang Zheng$^{\ddag}$,\ Yi Yang$^{\ddag}$,\ Qi Tian$^{\S}$,\ Shengjin Wang$^{\dag}\thanks{Corresponding Author}$ \\
 $^{\dag}$Tsinghua University\quad$^{\ddag}$University of Technology Sydney\quad $^{\S} $University of Texas at San Antonio\\
{\tt\small sunyf15@mails.tsinghua.edu.cn, \{liangzheng06, yee.i.yang\}@gmail.com }\\
{\tt\small Qi.tian@utsa.edu, wgsgj@tsinghua.edu.cn }
}
\maketitle

\begin{abstract}
Employing part-level features for pedestrian image description offers fine-grained information and has been verified as beneficial for person retrieval in very recent literature. A prerequisite of part discovery is that each part should be well located. Instead of using external cues, e.g.,  pose estimation, to directly locate parts, this paper lays emphasis on the content consistency within each part.

Specifically, we target at learning discriminative part-informed features for person retrieval and make two contributions. (i) A  network named Part-based Convolutional Baseline (PCB). Given an image input, it outputs a convolutional descriptor consisting of several part-level features. With a uniform partition strategy, PCB achieves competitive results with the state-of-the-art methods, 
proving itself as a strong convolutional baseline for person retrieval.
 (ii) A refined part pooling (RPP) method. Uniform partition inevitably incurs outliers in each part, which are in fact more similar to other parts. RPP re-assigns these outliers to the parts they are closest to, resulting in refined parts with enhanced within-part consistency. 
Experiment confirms that RPP allows PCB to gain another round of performance boost. For instance, on the Market-1501 dataset, we achieve (77.4+4.2)\% mAP and (92.3+1.5)\% rank-1 accuracy, surpassing the state of the art by a large margin.

\end{abstract}

\section{Introduction}\label{sec:introduction}
Person retrieval, also known as person re-identification (re-ID), aims at retrieving images of a specified pedestrian in a large database, given a query person-of-interest. Presently, deep learning methods dominate this community, with convincing superiority against hand-crafted competitors \cite{DBLP:journals/corr/ZhengYH16}. Deeply-learned representations provide high discriminative ability, especially when aggregated from deeply-learned part features. The latest state of the art on re-ID benchmarks are achieved with part-informed deep features \cite{Yao2017Deep,Su2017Pose,Zhao2017Deeply}.
\begin{figure}[t]
\setlength{\abovecaptionskip}{0cm}  
\setlength{\belowcaptionskip}{-1cm}  
\centering 
\includegraphics[width=0.95\linewidth]{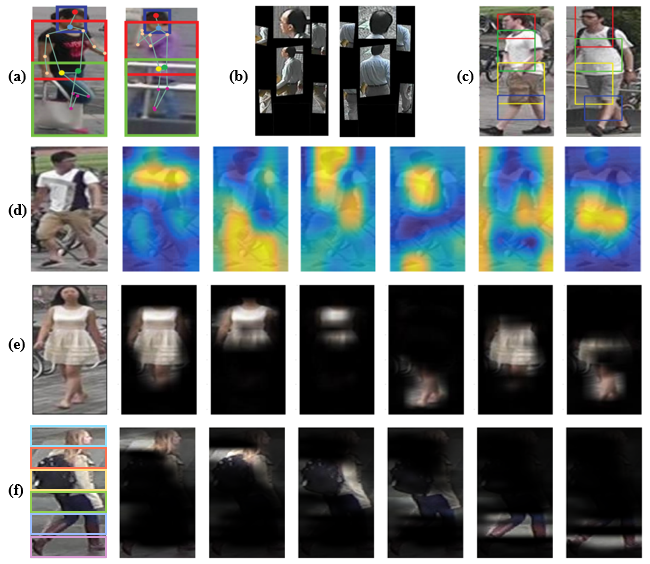}

   \caption{Partition strategies of several deep part models in person retrieval. (a) to (e): Partitioned parts by GLAD \cite{Wei2017GLAD}, PDC \cite{Su2017Pose}, DPL \cite{Yao2017Deep}, Hydra-plus \cite{Liu2017HydraPlus} and PAR \cite{Zhao2017Deeply}, respectively, which are cropped from the corresponding papers. (f): Our method employs a uniform partition and then refines each stripe. Both PAR \cite{Zhao2017Deeply} and our method conduct ``soft'' partition, but our method differs significantly from \cite{Zhao2017Deeply}, as detailed in Section \ref{sec:related}.}
\label{fig:part_cmp}
\end{figure}

An essential prerequisite of learning discriminative part features is that parts should be accurately located. Recent state-of-the-art methods vary on their partition strategies and can be divided into two groups accordingly. The first group \cite {DBLP:journals/corr/ZhengHLY17,Su2017Pose,Wei2017GLAD} leverage external cues, \emph{e.g.,} assistance from the latest progress  on human pose estimation \cite{pose:Long2015Fully,pose:CPM,pose:DeeperCut,pose:hourglass,pose:Cao2016Realtime}. They rely on external human pose estimation datasets and sophisticated pose estimator. The underlying datasets bias between pose estimation and person retrieval remains an obstacle against ideal semantic partition on person images. The other group \cite{Yao2017Deep,Zhao2017Deeply,Liu2017HydraPlus} abandon cues from semantic parts. They require no part labeling and yet achieve competitive accuracy with the first group. Some partition strategies are compared in Fig. \ref{fig:part_cmp}.
Against this background of progress on learning part-level deep features, we rethink the problem of what makes well-aligned parts. Semantic partitions may offer stable cues to good alignment but are prone to noisy pose detections. This paper, from another perspective, lays emphasis on the consistency within each part, which we speculate is vital to the spatial alignment. Then we arrive at our motivation that given coarsely partitioned parts, we aim to refine them to reinforce within-part consistency. Specifically, we make the following two contributions:

First, we propose a network named Part-based Convolutional Baseline (PCB) which conducts uniform partition on the conv-layer for learning part-level features. It does not explicitly partition the images. PCB takes a whole image as the input and outputs a convolutional feature. Being a classification net, the architecture of PCB is concise, with slight modifications on the backbone network. The training procedure is standard and requires no bells and whistles. We show that the convolutional descriptor has much higher discriminative ability than the commonly used fully-connected (FC) descriptor. On the Market-1501 dataset, for instance, the performance increases from 85.3\% rank-1 accuracy and 68.5\% mAP to 92.3\% (+7.0\%) rank-1 accuracy and 77.4\% (+8.9\%) mAP, surpassing many state-of-the-art methods by a large margin.

Second, we propose an adaptive pooling method to refine the uniform partition. We consider the motivation that within each part the contents should be consistent. We observe that under uniform partition, there exist outliers in each part. These outliers are, in fact, closer to contents in some other part, implying within-part inconsistency. Therefore, we refine the uniform partition by relocating those outliers to the part they are closest to, so that the within-part consistency is reinforced. An example of the refined parts is illustrated in Fig. \ref{fig:part_cmp}(f). With the proposed refined part pooling (RPP), performance on Market-1501 further increases to 93.8\% (+1.5\%) rank-1 accuracy and 81.6\% (+4.2\%) mAP.
In Section \ref{sec:network} and \ref{sec:pooling}, we describe the PCB and the refined part pooling, respectively.

In Section \ref{sec:experiment}, we combine the two methods, which achieves a new state of the art in person retrieval. Importantly, we demonstrate experimentally that the proposed refined parts are superior to attentive parts, \ie, parts learned with attention mechanisms.  

\section{Related Work}\label{sec:related}

\textbf{Hand-crafted part features for person retrieval.} Before deep learning methods dominated the re-ID research community, hand-crafted algorithms had developed approaches to learn part or local features. Gray and Tao \cite{Gray2008Viewpoint} partition pedestrians into horizontal stripes to extract color and texture features. Similar partitions have then been adopted by many works \cite{Engel2010Person,DBLP:journals/pami/ZhengGX13,Ma2014Domain,DBLP:conf/cvpr/LiaoHZL15}. Some other works employ more sophisticated strategy. Gheissari \etal \cite{Gheissari2006Person} divide the pedestrian into several triangles for part feature extraction. Cheng \etal \cite{cheng2011custom} employ pictorial structure to parse the pedestrian into semantic parts. Das \etal \cite{Das2014Consistent} apply HSV histograms on the head, torso and legs to capture spatial information.  

\textbf{Deeply-learned part features.} 
The state of the art on most person retrieval datasets is presently maintained by deep learning methods \cite{DBLP:journals/corr/ZhengYH16}. When learning part features for re-ID, the advantages of deep learning over hand-crafted algorithms are two-fold. First, deep features generically obtain stronger discriminative ability. Second, deep learning offers better tools for parsing pedestrians, which further benefits the part features. In particular, human pose estimation and landmark detection have achieved impressive progress \cite{pose:Long2015Fully,pose:hourglass,pose:Cao2016Realtime,pose:CPM,pose:DeeperCut}. Several recent works in re-ID employ these tools for pedestrian partition and report encouraging improvement \cite{DBLP:journals/corr/ZhengHLY17,Su2017Pose,Wei2017GLAD}. However, the underlying gap between datasets for pose estimation and person retrieval remains a problem when directly utilizing these pose estimation methods in an off-the-shelf manner. Others abandon the semantic cues for partition. Yao \etal \cite{Yao2017Deep} cluster the coordinates of max activations on feature maps to locate several regions of interest. Both Liu \etal \cite{Liu2017HydraPlus} and Zhao \etal \cite{Zhao2017Deeply} embed the attention mechanism \cite{Xu2015attention_mechanism} in the network, allowing the model to decide where to focus by itself.

\begin{figure*}[t]
\centering
\includegraphics[width=0.9\linewidth]{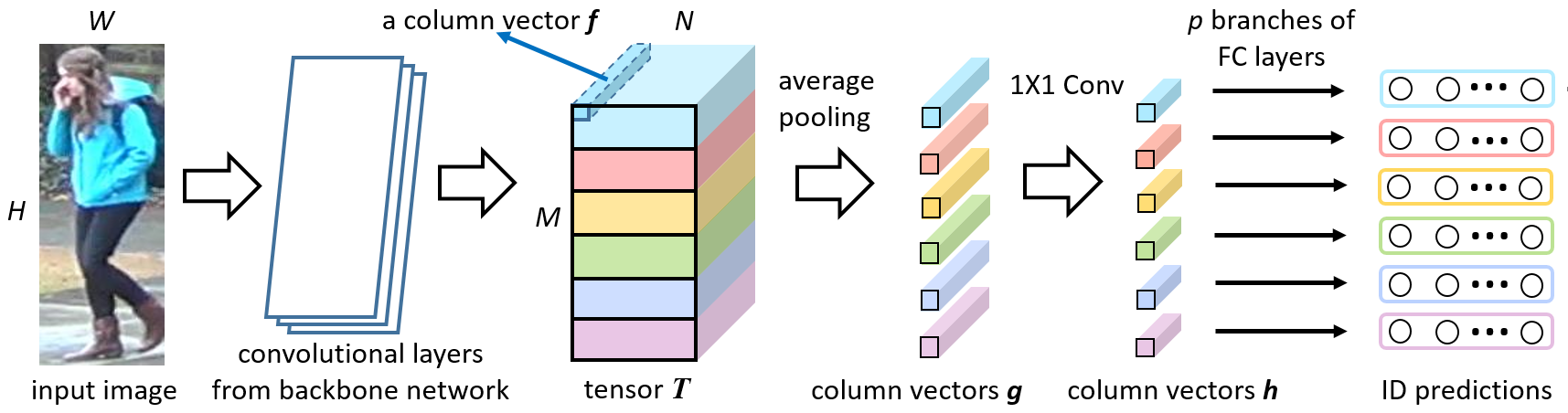}

\caption{Structure of PCB. The input image goes forward through the stacked convolutional layers from the backbone network to form a 3D tensor \bm{$T$}. PCB replaces the original global pooling layer with a conventional pooling layer, to spatially down-sample \bm{$T$} into $p$ pieces of column vectors \bm{$g$}. A following $1 \times 1$ kernel-sized convolutional layer reduces the dimension of \bm{$g$}. Finally, each dimension-reduced column vector \bm{$h$} is input into a classifier, respectively. Each classifier is implemented with a fully-connected (FC) layer and a sequential Softmax layer. During training, each classifier predicts the identity of the input image and is supervised by Cross-Entropy loss. During testing, either $p$ pieces of \bm{$g$} or \bm{$h$} are concatenated to form the final descriptor of the input image.}
\label{fig:structure}
\end{figure*}


\textbf{Deeply-learned part with attention mechanism.} A major contribution of this paper is the refined part pooling. We compare it with a recent work, PAR \cite{Yao2017Deep} by Zhao \etal in details. Both works employ a part-classifier to conduct ``soft'' partition on pedestrian images, as shown in Fig. \ref{fig:part_cmp}. Two works share the merit of requiring no part labeling for learning discriminative parts. However, the motivation, training methods, mechanism, and final performance of the two methods are quite different, to be detailed below.

\textbf{Motivation:} PAR aims at directly learning aligned parts while RPP aims to refine the pre-partitioned parts. \textbf{Working mechanism:} using attention method, PAR trains the part classifier in an unsupervised manner, while the training of RPP can be viewed as a semi-supervised process. \textbf{Training process:} RPP firstly trains an identity classification model with uniform partition and then utilizes the learned knowledge to induce the training of part classifier. \textbf{Performance:} the slightly more complicated training procedure rewards RPP with better interpretation and significantly higher performance. For instance on Market-1501, mAP achieved by PAR, PCB cooperating attention mechanism and the proposed RPP are 63.4\%, 74.6\% and 81.6\%, respectively. In addition, RPP has the potential to cooperate with various partition strategies. 

\section{PCB: A Strong Convolutional Baseline}\label{sec:network}
This section describes the structure of PCB and its comparison with several potential alternative structures.

\subsection{Structure of PCB}
\textbf{Backbone network.} PCB can take any network without hidden fully-connected layers designed for image classification as the backbone, \eg, Google Inception \cite{Szegedy2016Inception} and ResNet \cite{DBLP:conf/cvpr/HeZRS16}. 
This paper mainly employs ResNet50 with consideration of its competitive performance as well as its relatively concise architecture. 

\textbf{From backbone to PCB.} We reshape the backbone network to PCB with slight modifications, as illustrated in Fig. \ref{fig:structure}. The structure before the original global average pooling  (GAP) layer is maintained exactly the same as the backbone model. The difference is that the GAP layer and what follows are removed. When an images undergoes all the layers inherited from the backbone network, it becomes a 3D tensor \bm{$T$} of activations. In this paper, we define the vector of activations viewed along the channel axis as a \textbf{column vector}. 
Then, with a conventional average pooling, PCB partitions \bm{$T$} into $p$ horizontal stripes and averages all the column vectors in a same stripe into a single part-level column vector $\bm g_i$ ($i=1,2,\cdots, p$, the subscripts will be omitted unless necessary). 
Afterwards, PCB employs a convolutional layer to reduce the dimension of $\bm{g}$. According to our preliminary experiment, the dimension-reduced column vectors $\bm{h}$ are set to 256-dim. Finally, each $\bm{h}$ is input into a classifier, which is implemented with a fully-connected (FC) layer and a following Softmax function, to predict the identity (ID) of the input. 

During training, PCB is optimized by minimizing the sum of Cross-Entropy losses over $p$ pieces of ID predictions. During testing, either $p$ pieces of $\bm{g}$ or $\bm{h}$ are concatenated to form the final descriptor $\mathcal{G}$ or $\mathcal{H}$, \ie, $\mathcal{G} = [\bm{g}_1,\bm{g}_2,\cdots,\bm{g}_p]$ or $\mathcal{H} = [\bm{h}_1,\bm{h}_2,\cdots,\bm{h}_p]$. As observed in our experiment, employing $G$ achieves slightly higher accuracy, but at a larger computation cost, which is consistent with the observation in \cite{Sun2017SVDNet}. 

\subsection{Important Parameters.}\label{sec:params} PCB benefits from fine-grained spatial integration. Several key parameters, \emph{i.e.,} the input image size (\ie, \bm{$[H,W]}$), the spatial size of the tensor \bm{$T$} (\ie, \bm{$[M,N]}$), and the number of pooled column vectors (\ie, \bm{$p$}) are important to the performance of PCB. Note that \bm{$[M,N]}$ is determined by the spatial down-sampling rate of the backbone model, given the fixed-size input. Some deep object detection methods, \eg, SSD \cite{DBLP:conf/eccv/LiuAESRFB16} and R-FCN \cite{Dai2016R}, show that decreasing the down-sampling rate of the backbone network efficiently enriches the granularity of feature. PCB follows their success by removing the last spatial down-sampling operation in the backbone network to increase the size of $\bm{T}$. This manipulation considerably increases person retrieval accuracy with only very light computation cost added. The details can be accessed in Section \ref{sec:cmp_params}, which also provides insights to explain the phenomenon that partitioning tensor $\bm{T}$ into too many stripes (large $p$) compromises the discriminative ability of the learned feature.

Through our experiment, the optimized parameter settings for PCB are: 
\begin{itemize}
\item The input images are resized to $384 \times 128$, with a height to width ratio of 3:1. 
\item The spatial size of \bm{$T$} is set to $24 \times 8$.

\item \bm{$T$} is equally partitioned into $6$ horizontal stripes. 
\end{itemize}

\subsection{Potential Alternative Structures}\label{sec:alternative}

Given a same backbone network, there exist several potential alternative structures to learn part-level features. We enumerate two structures for comparison with PCB. 

\begin{itemize}
    \item Variant 1. Instead of making an ID prediction based on each $\bm{h}_i$ $(i=1,2,\cdots,p)$, it averages all $\bm{h}_i$ into a single vector \bm{$\overline{h}$}, which is then fully connected to an ID prediction vector. During testing, it also concatenates $\bm{g}$ or $\bm{h}$ to form the final descriptor. Variant 1 is featured by learning a convolutional descriptor under a single loss.
    \item Variant 2. It adopts exactly the same structure as PCB in Fig. \ref{fig:structure}. However, all the branches of FC classifiers in Variant 2 share a same set of parameters.
\end{itemize}

Both variants are experimentally validated as inferior to PCB. The superiority of PCB against Variant 1 shows that not only the convolutional descriptor itself, but also the respective supervision on each part, is vital for learning discriminative part-level features. The superiority of PCB against Variant 2 shows that sharing weights for classifiers, while reducing the risk of over-fitting, compromises the discriminative ability of the learned part-level features. The experiment details are to be viewed in Section \ref{sec:baseline}. 

\section{Refined Part Pooling}\label{sec:pooling}

Uniform partition for PCB is simple, effective, and yet to be improved. This section firstly explains the inconsistency phenomenon accompanying the uniform partition and then proposes the refined part pooling as a remedy to reinforce within-part consistency.

\subsection{Within-Part Inconsistency}\label{sec:inconsistency}
With focus on the tensor $\bm{T}$ to be spatially partitioned, our intuition of within-part inconsistency is: column vectors $f$ in a same part of $\bm{T}$ should be similar to each other and be dissimilar to column vectors in other parts; otherwise the phenomenon of within-part inconsistency occurs, implying that the parts are partitioned inappropriately.

\begin{figure}[t]
\setlength{\abovecaptionskip}{-0.1cm} 
\setlength{\belowcaptionskip}{-0.2cm}
\begin{center}
\includegraphics[width=0.9\linewidth]{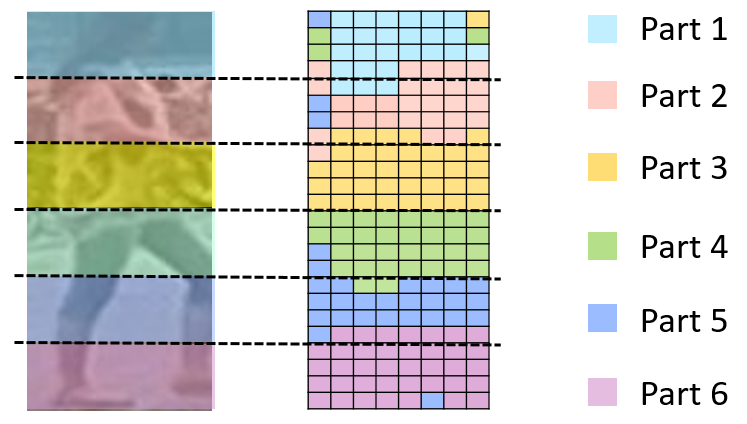}
\end{center}
   \caption{Visualization of within-part inconsistency. \bm{$T$}. Left: \bm{$T$} is equally partitioned to $p=6$ horizontal stripes (parts) during training. Right: Every column vector in \bm{$T$} is denoted with a small rectangle and painted in the color of its closest part.}
\label{fig:outlier}
\end{figure}

After training PCB to convergence, we compare the similarities between each $f$ and $\bm g_i$ ($i=1,2,\cdots,p$), \ie, the average-pooled column vector of each stripe, by measuring cosine distance. If $f$ is closest to $\bm g_i$, $f$ is inferred as closest to the $i$th part, correspondingly. By doing this, we find the closest part to each $f$, as exampled in Fig. \ref{fig:outlier}. Each column vector is denoted by a small rectangle and painted in the color of its closest part. 

Two phenomena are observed. First, most column vectors in a same horizontal stripe are clustered together (though there are no explicit constraints for this effect). Second, there exist many outliers, while designated to a specified horizontal stripe (part) during training, which are more similar to another part. The existence of these outliers suggests that they are inherently more consistent with column vectors in another part.  

\subsection{Relocating Outliers}\label{sec:relocate}
We propose the refined part pooling to correct within-part inconsistency. Our goal is to assign all the column vectors according to their similarities to each part, so that the outliers will be relocated.  

To this end, we need to classify all the column vectors $f$ in \bm{$T$} on the fly. Based on the already-learned \bm{$T$}, we use a linear layer followed by Softmax activation as a part classifier as follows:
\begin{equation}\label{eq:classifier} 
P(P_i|f)=softmax(W_i^Tf)=\frac{\exp(W_i^Tf)}{\sum\limits_{j=1}^p\exp(W_j^Tf)},
\end{equation}
where $P(P_i|f)$ is the predicted probability of $f$ belonging to part $P_i$, $p$ is the number of pre-defined parts (\ie, $p=6$ in PCB), and $W$ is the trainable weight matrix of the part classifier, whose training procedure is to be detailed in Section \ref{sec:train}. 

Given a column vector $f$ in $\bm T$ and the predicted probability of $f$ belonging to part $P_i$, we assign $f$ to part $P_i$ with $P(P_i|f)$ as the confidence. Correspondingly, each part $P_i(i=1,2,...,p)$ is sampled from all column vectors $f$ with $P(P_i|f)$ as the sampling weight, \ie,

\begin{equation}P_i=\{P(P_i|f) \times f, \forall {f \in F}\},\end{equation}
where $F$ is the complete set of column vectors in tensor \bm{$T$}, $\{\bullet\}$ denotes the sampling operation to form an aggregate. 

\begin{figure}[t]
\setlength{\abovecaptionskip}{-0.1cm} 
\setlength{\belowcaptionskip}{-0.2cm}
\begin{center}
\includegraphics[width=1.0\linewidth]{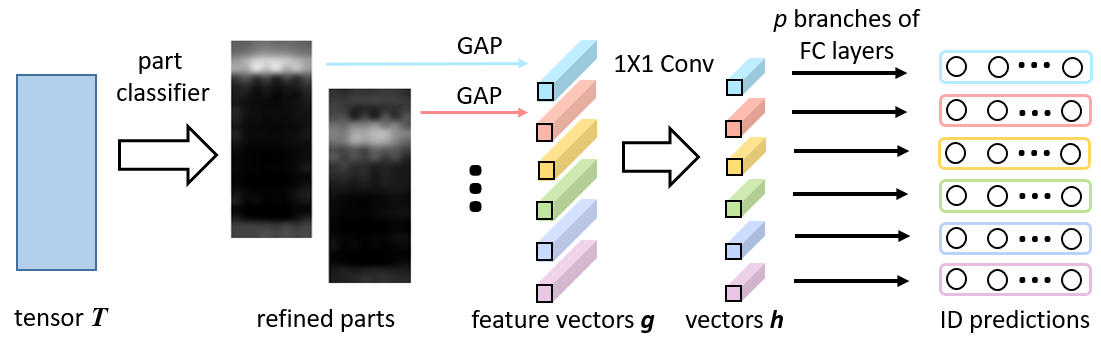}
\end{center}
   \caption{PCB in combination with refined part pooling. The 3D tensor \bm{$T$} is denoted simply by a rectangle instead of a cube as we focus on the spatial partition. Layers before \bm{$T$} are omitted as they remain unchanged compared with Fig. \ref{fig:structure}. A part classifier predicts the probability of each column vector belonging to $p$ parts. Then each part is sampled from all the column vectors with the corresponding probability as the sampling weight. GAP denotes global average pooling. }
\label{fig:reshape}
\end{figure}

By doing this, the proposed refined part pooling conducts a ``soft" and adaptive partition to refine the original ``hard" and uniform partition, and the outliers originated from the uniform partition will be relocated. In combination with refined part pooling described above, PCB is further reshaped into Fig. \ref{fig:reshape}. Refined part pooling, \ie, the part classifier along with the following sampling operation, replaces the original average pooling. The structure of all the other layers remain exactly the same as in Fig. \ref{fig:structure}.

\subsection{Induced Training for Part Classifier}\label{sec:train}

There lacks explicit supervisory information for learning $W$ of the part classifier in Eq. \ref{eq:classifier}. 
We design an induced training procedure instead, as illustrated in Alg. \ref{alg:induction}.

\begin{itemize}
\item First, a standard PCB model is trained to convergence with $\bm T$ equally partitioned. 
\item Second, we remove the original average pooling layer after $\bm T$ and append a $p$-category part classifier on $\bm T$. New parts are sampled from $\bm T$ according to the prediction of the part classifier, as detailed in Section \ref{sec:relocate}.
\item Third, we set the all the already learned layers in PCB fixed, leaving only the part classifier trainable. Then we retrain the model on training set. In this condition, the model still expects the tensor $\bm T$ to be equally partitioned, otherwise it will predict incorrect about the identities of training images. So Step 3 penalizes the part classifier until it conducts partition close to the original uniform partition, whereas the part classifier is prone to categorize inherently similar column vectors into a same part. A state of balance will be reached as a result of Step 3. 
\item Finally, all the layers are allowed to be updated. The whole net, \ie, PCB along with the part classifier are fine-tuned for overall optimization.
\end{itemize}

In the above training procedure, PCB model trained in Step1 induces the training of the part classifier. Step3 and 4 converges very fast, requiring 10 more epochs in total.

\begin{algorithm}
\caption{Induced training for part classifier}
\textbf{Step 1.} A standard PCB is trained to convergence with uniform partition.\\
\textbf{Step 2.} A $p$-category part classifier is appended on the tensor $\bm T$. \\
\textbf{Step 3.} All the pre-trained layers of PCB are fixed. Only the part classifier is trainable. The model is trained until convergence again.\\
\textbf{Step 4.} The whole net is fine-tuned to convergence for overall optimization.\\
\label{alg:induction}
\end{algorithm}

\subsection{Discussions on Refined Part Pooling}

With step 1 in Alg. \ref{alg:induction} skipped, the training can also converge. In this case, the training will be similar to PAR \cite{Zhao2017Deeply} which employs attention mechanism to align parts, as introduced in Section \ref{sec:related}. We compare both approaches, \ie, training part classifier with or without step 1, in experiments and find out that the induction procedure matters. Without the proposed induction, the performance turns out significantly lower. For example on Market-1501, when induction is applied, PCB in combination with refined part pooling achieves 80.9\% mAP. When induction is removed, mAP decreases to 74.6\%. It implies that the proposed induced training is superior to attention mechanism on PCB. The details can be accessed in Section \ref{sec:cmp_attention}. 

\setlength{\tabcolsep}{5.6pt}
\begin{table*}[]
\setlength{\abovecaptionskip}{-0.1cm}
\setlength{\belowcaptionskip}{-0.2cm}
\begin{center}
\begin{tabular}{l|c|c|c c c c|c c c c|c c c c}
\hline
\multicolumn{1}{l|}{\multirow{2}{*}{Models }}&\multicolumn{1}{c|}{\multirow{2}{*}{Feature}}&\multicolumn{1}{c|}{\multirow{2}{*}{dim}}&\multicolumn{4}{c|}{Market-1501} & \multicolumn{4}{c|}{DukeMTMC-reID} & \multicolumn{4}{c}{CUHK03}\\ 
\cline{4-15}
\multicolumn{1}{c|}{}&\multicolumn{1}{c|}{}&&\multicolumn{1}{c}{R-1}&{R-5}&{R-10}&{mAP}&{R-1}&{R-5}&{R-10}&{mAP}&{R-1}&{R-5}&{R-10}&{mAP}\\
\hline
IDE& pool5 & 2048 &85.3&94.0&96.3&68.5  &73.2&84.0&87.6& 52.8  &43.8&62.7&71.2&38.9\\
IDE& FC & 256 &83.8&93.1&95.8&67.7  &72.4& 83.0&87.1& 51.6   &43.3&62.5&71.0&38.3\\ 

Variant 1& $\mathcal{G}$ &12288 &86.7&95.2&96.5&69.4 &73.9&84.6&88.1& 53.2 &43.6&62.9&71.3&38.8\\
Variant 1& $\mathcal{H}$ &1536 &85.6&94.3&96.3&68.3 &72.8&83.3&87.2& 52.5 &44.1&63.0&71.5&39.1\\

Variant 2 &$\mathcal{G}$ &12288 &91.2&96.6&97.7&75.0 &80.2&88.8&91.3&62.8    &52.6&72.4&80.9 &45.8\\
Variant 2 &$\mathcal{H}$ &1536 &91.0&96.6&97.6&75.3 &80.0&88.1&90.4&62.6     &54.0&73.7&81.4&47.2\\

\hline
PCB &$\mathcal{G}$ &12288    &92.3&97.2&98.2&77.4    &81.7&89.7&91.9&66.1    &59.7&77.7&85.2&53.2\\
PCB &$\mathcal{H}$ &1536    &92.4&97.0&97.9&77.3   &81.9&89.4&91.6&65.3     &61.3&78.6&85.6&54.2\\

PCB+RPP& $\mathcal{G}$ & 12288   &\textbf{93.8}&\textbf{97.5}&\textbf{98.5}&\textbf{81.6}    &\textbf{83.3}&\textbf{90.5}&\textbf{92.5}&\textbf{69.2}     &62.8&79.8&86.8&56.7\\
PCB+RPP &$\mathcal{H}$ & 1536  &93.1&97.4&98.3&81.0   &82.9&90.1&92.3&68.5      &\textbf{63.7}&\textbf{80.6}&\textbf{86.9}&\textbf{57.5}\\

\hline 
\end{tabular}
\end{center}
\caption{Comparison of the proposed method with IDE and 2 variants. Both variants are described in Section \ref{sec:alternative}. pool5: output of Pool5 layer in ResNet50. FC: output of the appended FC layer for dimension reduction.  $\mathcal{G}$ ($\mathcal{H}$): feature representation assembled with column vectors $\bm{g}$ ($\bm{h}$). Both $\bm{g}$ and $\bm{h}$ are illustrated in Fig. \ref{fig:structure}.}
\label{table:cmpbasl}
\end{table*}


\section{Experiments}\label{sec:experiment}
\subsection{Datasets and Settings}
\textbf{Datasets.} This paper uses three datasets for evaluation, \ie, \textbf{Market-1501} \cite{DBLP:conf/iccv/ZhengSTWWT15}, \textbf{DukeMTMC-reID} \cite{ristani2016MTMC,zheng2017unlabeled}, and \textbf{CUHK03} \cite{DBLP:conf/cvpr/LiZXW14}. The Market-1501 dataset contains 1,501 identities observed under 6 camera viewpoints, 19,732 gallery images and 12,936 training images detected by DPM \cite{felzenszwalb2008discriminatively}. The DukeMTMC-reID dataset contains 1,404 identities, 16,522 training images, 2,228 queries, and 17,661 gallery images. With so many images captured by 8 cameras, DukeMTMC-reID manifests itself as one of the most challenging re-ID datasets up to now. The CUHK03 dataset contains 13,164 images of 1,467 identities. Each identity is observed by 2 cameras. CUHK03 offers both hand-labeled and DPM-detected bounding boxes, and we use the latter in this paper. CUHK03 originally adopts 20 random train/test splits, which is time-consuming for deep learning. So we adopt the new training/testing protocol proposed in \cite{DBLP:conf/cvpr/ZhongZCL17}. For Market-1501 and DukeMTMC-reID, we use the evaluation packages provided by \cite{DBLP:conf/iccv/ZhengSTWWT15} and \cite{zheng2017unlabeled}, respectively.
All the experiment evaluates the single-query setting. 
Moreover, for simplicity we do not use re-ranking algorithms which considerably improve mAP \cite{DBLP:conf/cvpr/ZhongZCL17}. Our results are compared with reported results without re-ranking. 

\subsection{Implementation details} 
\textbf{Implementation of IDE for comparison.} We note that the IDE model specified in \cite{DBLP:journals/corr/ZhengYH16} is a commonly used baseline in deep re-ID systems \cite{DBLP:journals/corr/ZhengYH16,DBLP:journals/corr/ZhengHLY17,DBLP:conf/cvpr/XiaoLOW16,geng2016deep,Sun2017SVDNet,Zheng2017PAN,zheng2017unlabeled,Zhong2017Random}. In contrast to the proposed PCB, the IDE model learns a global descriptor. For comparison, we implement the IDE model on the same backbone network, \ie, ResNet50, and with several optimizations over the original one in \cite{DBLP:journals/corr/ZhengYH16}, as follows.
1) After the ``pool5" layer in ResNet50, we append a fully-connected layer followed by Batch Normalization and ReLU. The output dimension of the appended FC layer is set to 256-dim. 2) We apply dropout on ``pool5" layer. Although there are no trainable parameters in ``pool5" layer, there is evidence that applying Dropout on it, which outputs a high dimensional feature vector of 2048d, effectively avoids over-fitting and gains considerable improvement \cite{Zheng2017PAN,zheng2017unlabeled}. We empirically set the dropout ratio to 0.5.
On Market-1501, our implemented IDE achieves 85.3\% rank-1 accuracy and 68.5\% mAP, which is a bit higher than the implementation in \cite{Zhong2017Random}. 

\textbf{Training.} The training images are augmented with horizontal flip and normalization. We set batch size to 64 and train the model for 60 epochs with base learning rate initialized at 0.1 and decayed to 0.01 after 40 epochs. The backbone model is pre-trained on ImageNet \cite{Deng2009imagenet}. The learning rate for all the pre-trained layers are set to  $0.1 \times$ of the base learning rate. When employing refined part pooling for boosting, we append another 10 epochs with learning rate set to 0.01. With two NVIDIA TITAN XP GPUs and Pytorch as the platform, training an IDE model and a standard PCB on Market-1501 (12,936 training images) consumes about 40 and 50 minutes, respectively. The increased training time of PCB is mainly caused by the cancellation of the last spatial down-sample operation in the Conv5 layer, which enlarges the tensor \bm{$T$} by $4 \times$.

\subsection{Performance evaluation}\label{sec:baseline}
We evaluate our method on three datasets, with results shown in Table \ref{table:cmpbasl}. Both uniform partition (PCB) and refined part pooling (PCB+RPP) are tested.

\textbf{PCB is a strong baseline.}
Comparing PCB and IDE, the prior commonly used baseline in many works \cite{DBLP:journals/corr/ZhengYH16,DBLP:journals/corr/ZhengHLY17,DBLP:conf/cvpr/XiaoLOW16,geng2016deep,Sun2017SVDNet,Zheng2017PAN,zheng2017unlabeled,Zhong2017Random},
we clearly observe the significant advantage of PCB: mAP on three datasets increases from 68.5\%, 52.8\% and 38.9\% to 77.4\% (+8.9\%), 66.1\% (+13.3\%) and 54.2\% (+15.3\%), respectively. This indicates that integrating part information increases the discriminative ability of the feature. The structure of PCB is as concise as that of IDE, and training PCB requires nothing more than training a canonical classification network. We hope it will serve as a baseline for person retrieval task.

\textbf{Refined part pooling (RPP) improves PCB especially in mAP.} From Table \ref{table:cmpbasl}, while PCB already has a high accuracy, RPP brings further improvement to it. On the three datasets, the improvement in rank-1 accuracy is +1.5\%, +1.6\%, and +3.1\%, respectively; the improvement in mAP is +4.2\%, +3.1\%, and +3.5\%, respectively. The improvement is larger in mAP than in rank-1 accuracy. In fact, rank-1 accuracy characterizes the ability to retrieve the easiest match in the camera network, while mAP indicates the ability to find all the matches. So the results indicate that RPP is especially beneficial in finding more challenging matches. 

\textbf{The benefit of using $p$ losses.} To validate the usage of $p$ branches of losses in Fig. \ref{fig:structure}, we compare our method with Variant 1 which learns the convolutional descriptor under a single classification loss. Table \ref{table:cmpbasl} suggests that Variant 1 yields much lower accuracy than PCB, implying that employing a respective loss for each part is vital for learning discriminative part features. 

\textbf{The benefit of NOT sharing parameters among identity classifiers.} In Fig. \ref{fig:structure}, PCB inputs each column vector $\bm h$ to a FC layer before the Softmax loss. We compare our proposal (not sharing FC layer parameters) with Variant 2 (sharing FC layer parameters). From Table \ref{table:cmpbasl}, PCB is higher than Variant 2 by 2.4\%, 3.3\%, and 7.4\% on the three datasets, respectively. This suggests that sharing parameters among the final FC layers is inferior.

\setlength{\tabcolsep}{8.3pt}
\begin{table}
\renewcommand\arraystretch{1.1}
\begin{center}
\begin{tabular}{l|cccc}
\hline
Methods &R-1&R-5&R-10&mAP \\
\hline
BoW+kissme \cite{DBLP:conf/iccv/ZhengSTWWT15}  &44.4	&63.9	&72.2  &20.8\\
WARCA\cite{Jose2016Scalable_WARCA}    &45.2	&68.1	&76.0  &-\\
KLFDA\cite{Karanam2016A_KLFDA}  &46.5	&71.1	&79.9  &-\\
\hline
SOMAnet\cite{barbosa2017looking} &73.9&-&-&47.9\\
SVDNet\cite{Sun2017SVDNet} &{82.3}&{92.3}&{95.2}&{62.1}\\
PAN\cite{Zheng2017PAN} &82.8&-&-&63.4\\
Transfer \cite{geng2016deep}  &{83.7}&-&-&{65.5}\\
Triplet Loss \cite{Hermans2017DefenseTriplet}  &84.9	&94.2  &-   &69.1\\
DML  \cite{Zhang2017DeepMutualLearning}      &87.7&-&-&68.8\\

\hline
MultiRegion \cite{Ustinova2015Multiregion}  &66.4	&85.0	&90.2   &41.2\\
HydraPlus \cite{Liu2017HydraPlus} &76.9&91.3&94.5&-\\
PAR \cite{Zhao2017Deeply}   &81.0&92.0&94.7&63.4\\
MultiLoss \cite{Li2017_multiloss} &83.9&-&-&64.4\\
PDC* \cite{Su2017Pose}    &84.4&92.7&94.9&63.4\\
PartLoss \cite{Yao2017Deep}  &88.2&-&-&69.3\\
MultiScale \cite{ChenPerson_multiscale}&88.9&-&-&73.1\\
GLAD* \cite{Wei2017GLAD}         &89.9&-&-&73.9\\

\hline
PCB  &92.3&97.2&98.2&77.4\\
PCB+RPP &\textbf{93.8}&\textbf{97.5}&\textbf{98.5}&\textbf{81.6} \\

\hline
\end{tabular}
\end{center}
\setlength{\abovecaptionskip}{0cm}
\caption{Comparison of the proposed method with the art on Market-1501. The compared methods are categorized into 3 groups. Group 1: hand-crafted methods. Group 2: deep learning methods employing global feature. Group 3: deep learning methods employing part features. * denotes those requiring auxiliary part labels. Our method is denoted by ``PCB'' and ``PCB+RPP''.}
\label{table:cmp_sota_market}
\end{table}


\textbf{Comparison with state of the art.}
We compare PCB and PCB+RPP with state of the art. Comparisons on Market-1501 are detailed in Table \ref{table:cmp_sota_market}. The compared methods are categorized into three groups, \ie, hand-crafted methods, deep learning methods with global feature and deep learning methods with part features. Relying on uniform partition only, PCB surpasses all the prior methods, including \cite{Su2017Pose,Wei2017GLAD} which require auxiliary part labeling to deliberately align parts. The performance lead is further enlarged by the proposed refined part pooling. 

Comparisons on DukeMTMC-reID and CUHK03 (new training/testing protocol) are summarized in Table \ref{table:duke}. In the compared methods,  PCB exceeds \cite{ChenPerson_multiscale} by +5.5\% and 17.2\% in mAP on the two datasets, respectively. PCB+RPP (refined part pooling) further surpasses it by a large margin of +8.6\% mAP on DukeMTMC-reID and +20.5\% mAP on CUHK03. PCB+RPP yields higher accuracy than ``TriNet+Era'' and ``SVDNet+Era'' \cite{Zhong2017Random} which are enhanced by extra data augmentation. 

In this paper, \textbf{we report mAP = 81.6\%, 69.2\%, 57.5\% and Rank-1 = 93.8\%, 83.3\% and 63.7\% for Market-1501, Duke and CUHK03}, respectively, setting new state of the art on the three datasets. All the results are achieved under the single-query mode without re-ranking. Re-ranking methods will further boost the performance especially mAP. For example, when ``PCB+RPP" is combined with the method in \cite{DBLP:conf/cvpr/ZhongZCL17}, mAP and Rank-1 accuracy on Market-1501 increases to \textbf{91.9\%} and \textbf{95.1\%}, respectively.

\setlength{\tabcolsep}{4.6pt}
\begin{table}
\begin{center}
\begin{tabular}{l|cc|cc}
\hline
\multicolumn{1}{l|}{\multirow{2}{*}{Methods}}&\multicolumn{2}{c|}{DukeMTMC-reID}&\multicolumn{2}{c}{CUHK03}\\
\cline{2-5}
\multicolumn{1}{c|}{}&rank-1&mAP&rank-1&mAP\\
\hline
BoW+kissme \cite{DBLP:conf/iccv/ZhengSTWWT15} & 25.1 & 12.2 & 6.4 & 6.4\\
LOMO+XQDA \cite{DBLP:conf/cvpr/LiaoHZL15} & 30.8 & 17.0 & 12.8 & 11.5 \\
GAN \cite{zheng2017unlabeled} & {67.7}  & {47.1} & - & - \\
PAN \cite{Zheng2017PAN} & 71.6& 51.5 & 36.3 &34.0\\
SVDNet \cite{Sun2017SVDNet} &76.7&56.8&41.5&37.3\\
MultiScale \cite{ChenPerson_multiscale}   &79.2&60.6&40.7&37.0\\
TriNet+Era \cite{Zhong2017Random} &73.0&56.6&55.5&50.7\\
SVDNet+Era \cite{Zhong2017Random} &79.3&62.4&48.7&43.5\\
\hline
PCB (UP) &81.8&66.1 &61.3 & 54.2 \\
PCB (RPP)&\textbf{83.3}& \textbf{69.2} & \textbf{63.7} & \textbf{57.5} \\

\hline
\end{tabular}
\end{center}
\setlength{\abovecaptionskip}{0cm} 
\setlength{\belowcaptionskip}{0pt} 
\caption{Comparison with prior art on DukeMTMC-reID and CUHK03. Rank-1 accuracy (\%) and mAP (\%) are shown. }
\label{table:duke}
\end{table}


\subsection{Parameters Analysis}\label{sec:cmp_params}
We analyze some important parameters of PCB (and with RPP) introduced in Section \ref{sec:params} on Market-1501. Once optimized, the same parameters are used for all the three datasets. 

\begin{figure}[t]
\begin{center}
\includegraphics[width=1.0\linewidth]{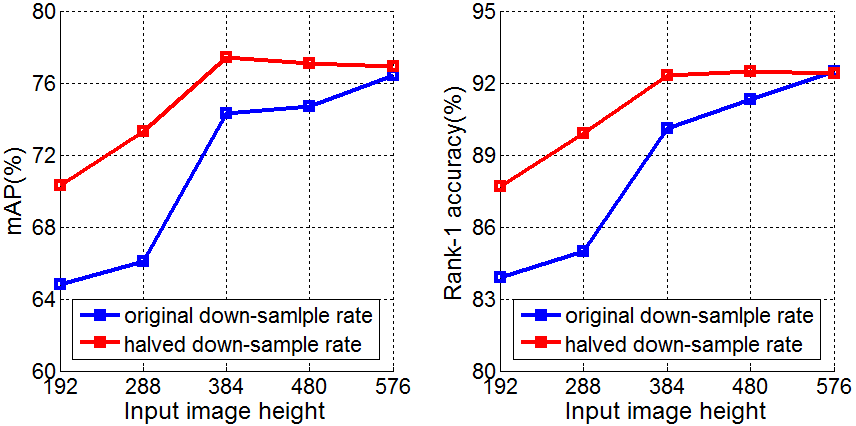}
\end{center}

   \caption{Impact of the size of the image and \bm{$T$}. Rank-1 accuracy and mAP are compared. Using the original and halved down-sampling rates, two different sizes of \bm${T}$ are compared.} 
\label{fig:cmp_sizes}
\end{figure}

\textbf{The size of images and tensor \bm{$T$}.} We vary the image size from $192 \times 64$ to $576 \times 192$, using $96 \times 32$ as interval. Two down-sampling rates are tested, \ie, the original rate, and a halved rate (larger \bm{$T$}).   
We exhaustively train all these models on PCB and report their performance in Fig. \ref{fig:cmp_sizes}. Two phenomena are observed.

First, a larger image size benefits the learned part feature. Both mAP and rank-1 accuracy increase with the image size until reaching a stable performance.

Second, a smaller down-sampling rate, \ie, a larger spatial size of tensor \bm{$T$} enhances the performance, especially when using relatively small images as input. In Fig. \ref{fig:cmp_sizes},  PCB  using $384 \times 128$ input and halved down-sampling rate achieves almost the same performance as PCB using $576 \times 192$ input and the original down-sampling rate. We recommend the manipulation of halving the down-sampling rate with consideration of the computing efficiency.

\begin{figure}[t]
\setlength{\belowcaptionskip}{-0.4cm}
\begin{center}
\includegraphics[width=1.0\linewidth]{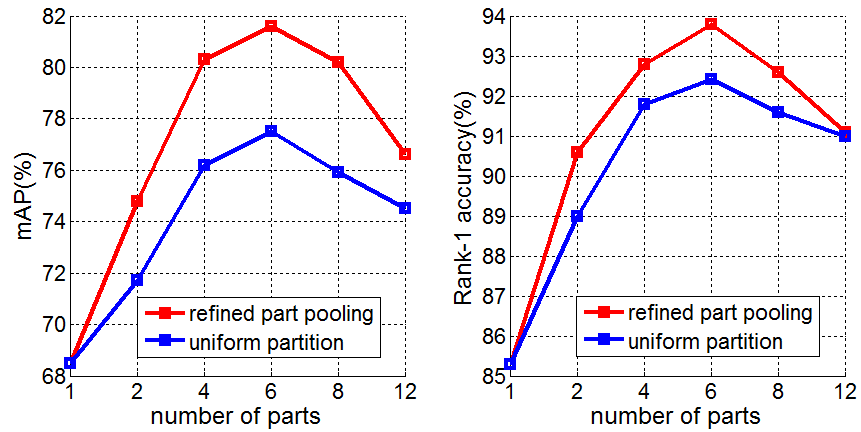}
\end{center}
   \caption{Impact of $p$. Rank-1 accuracy and mAP are compared. We compare PCB both with and without  the refined part pooling.}
\label{fig:stripes}
\end{figure}

\begin{figure}[t]

\begin{center}
\includegraphics[width=1.0\linewidth]{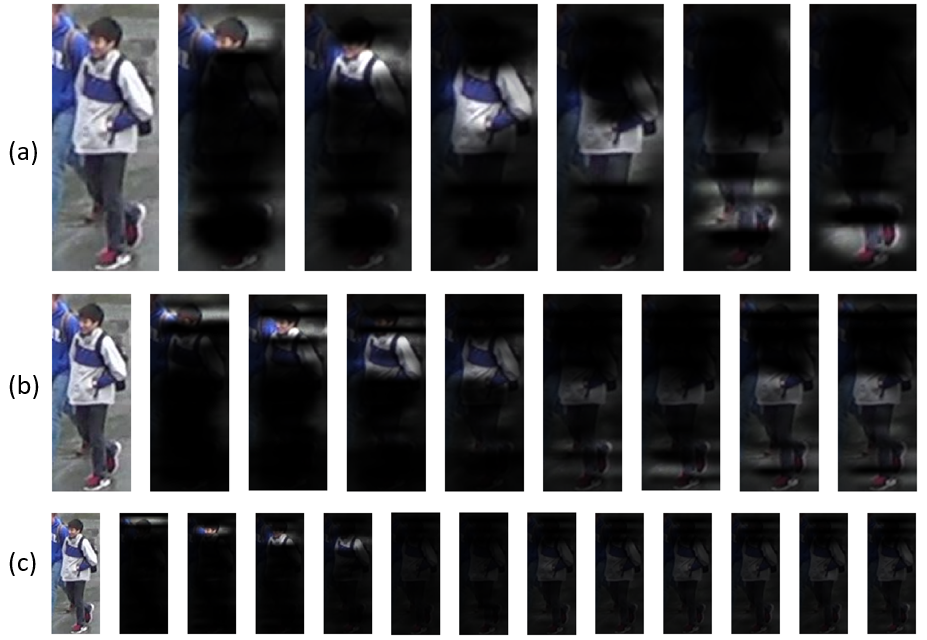}
\end{center}
   \caption{Visualization of the refined parts under different $p$ values. When $p$ = 8 or 12, some parts repeat with others or become empty.}
\label{fig:vis_stripes}
\end{figure}

\textbf{The number of parts $p$.} Intuitively, $p$ determines the granularity of the part feature. When $p$=1, the learned feature is a global one. As $p$ increases, retrieval accuracy improves at first. However, accuracy does not always increase with $p$, as illustrated in Fig. \ref{fig:stripes}. When $p = 8$ or 12, the performance drops dramatically, regardless of using refined part pooling. A visualization of the refined parts offers insights into this phenomenon, as illustrated in Fig. \ref{fig:vis_stripes}. When $p$ increases to 8 or 12, some of the refined parts are very similar to others and some may collapse to an empty part. As a result, an over-increased $p$ actually compromises the discriminative ability of the part features. In real-world applications, we would recommend to use $p=6$ parts.

\subsection{Induction and Attention Mechanism}\label{sec:cmp_attention}
In this work, when training the part classifier in Alg. \ref{alg:induction}, a PCB pre-trained with uniform partition is required. The knowledge learned under uniform partition induces the subsequent training of the part classifier. \emph{Without PCB pre-training, the network learns to partition \bm{$T$} under no induction and becomes similar to methods driven by attention mechanism.} We conduct an ablation experiment on Market-1501 and DukeMTMC-reID to compare the two approaches. Results are presented in Table \ref{tab:cmp_with_attention}, from which three observations can be drawn.

First, no matter which partition strategy is applied in PCB, it significantly outperforms PAR \cite{Zhao2017Deeply}, which learns to partition through attention mechanism. Second, the attention mechanism also works based on the structure of PCB. Under the ``RPP (w/o induction)" setting, the network learns to focus on several parts through attention mechanism, and achieves substantial improvement over IDE, which learns a global descriptor. Third, the induction procedure (PCB training) is critical. When the part classifier is trained without induction, the retrieval performance drops dramatically, compared with the performance achieved by ``PCB+RPP". It implies that the refined parts learned through induction is superior to the parts learned through attention mechanism.
Partitioned results with induction and attention mechanism are visualized in Fig. \ref{fig:part_cmp}.
\setlength{\tabcolsep}{2.9pt}
\begin{table}
\begin{center}
\begin{tabular}{l|cc|cc}
\hline
\multicolumn{1}{l|}{\multirow{2}{*}{Methods}}&\multicolumn{2}{c|}{Market-1501}&\multicolumn{2}{c}{DukeMTMC}\\
\cline{2-5}
\multicolumn{1}{c|}{}&rank-1&mAP&rank-1&mAP\\
\hline
PAR\cite{Zhao2017Deeply}   &81.0&63.4   &-&-\\
IDE   &85.3 & 68.5   & 73.2 &52.8\\

RPP (w/o induction) &88.7&74.6  &78.8&60.9 \\

PCB &92.3&77.4  &81.7&66.1 \\

PCB+RPP  &93.8&81.6  &83.3& 69.2 \\
\hline
\end{tabular}
\end{center}
\setlength{\abovecaptionskip}{0cm}
\caption{Ablation study of induction on Market-1501. PAR learns to focus on several parts to discriminate person with attention mechanisms. RPP (w/o induction) means no induction for learning the refined parts and the network learns to focus on several parts with attention mechanism.}
\label{tab:cmp_with_attention}
\end{table}

\section{Conclusion}
This paper makes two contributions to solving the pedestrian retrieval problem. First, we propose a Part-based Convolutional Baseline (PCB) for learning part-informed features. PCB employs a simple uniform partition strategy and assembles part-informed features into a convolutional descriptor. PCB advances the state of the art to a new level, proving itself as a strong baseline for learning part-informed features. 
Despite the fact that PCB with uniform partition is simple and effective, it is yet to be improved. We propose the refined part pooling to reinforce the within-part consistency in each part. After refinement, similar column vectors are concluded into a same part, making each part more internally consistent. Refined part pooling requires no part labeling information and improves PCB considerably.


{\small
\bibliographystyle{ieee}
\bibliography{egbib}
}

\end{document}